\documentclass[final]{cvpr}

\usepackage{times}
\usepackage{epsfig}
\usepackage{graphicx}
\usepackage{amsmath}
\usepackage{amssymb}
\usepackage{caption}
\usepackage[pagebackref=true,breaklinks=true,colorlinks,bookmarks=false]{hyperref}
\usepackage{unitsdef}
\usepackage{tabu}
\usepackage{multirow}
\def\cvprPaperID{5125} % *** Enter the CVPR Paper ID here
\def\confYear{CVPR 2021}

\begin{document}

\title{Shot Contrastive Self-Supervised Learning for Scene Boundary Detection \vspace{-0.45cm}}

\author{Shixing Chen \quad Xiaohan Nie\thanks{Equal contribution.} \qquad David Fan\footnotemark[1] \qquad Dongqing Zhang \quad Vimal Bhat \quad Raffay Hamid\\
	Amazon Prime Video\\
	{\tt\small \{shixic, nxiaohan, fandavi, zdongqin, vimalb, raffay\}@amazon.com}
}

\teaser{
	\vspace{0.01cm}
	\includegraphics[width=0.8\textwidth]{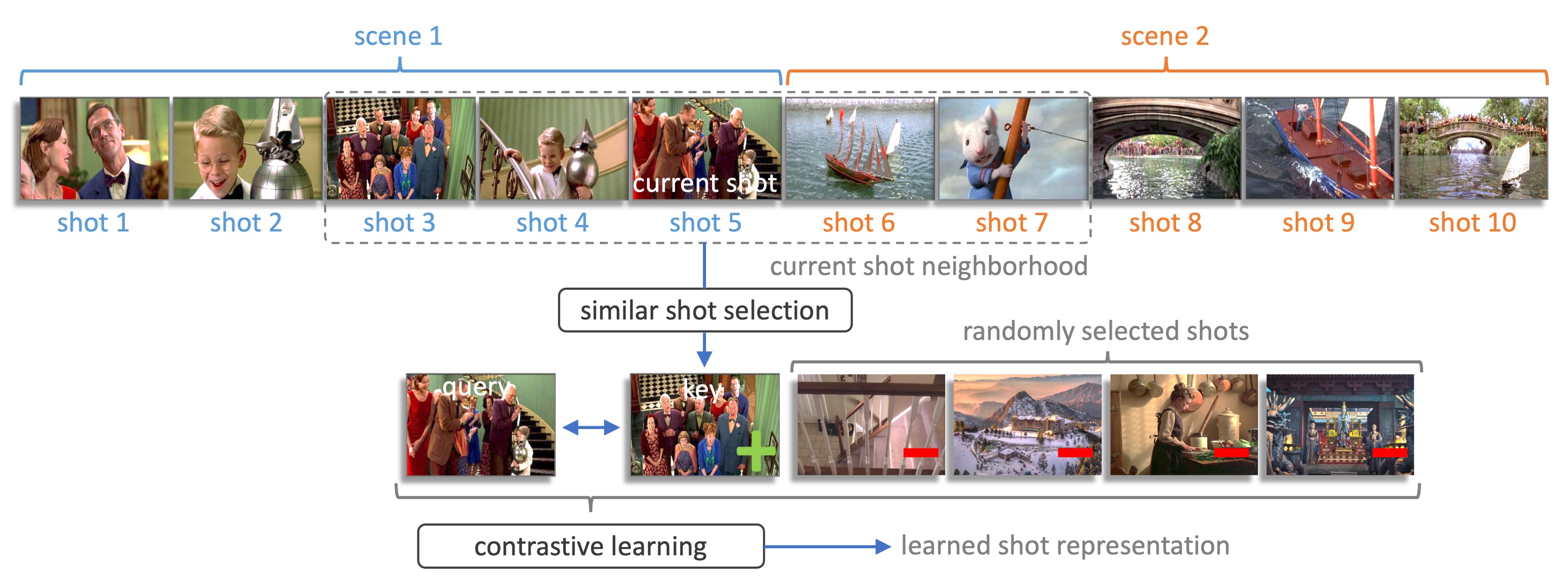}
	\vspace{-0.1cm}\captionof{figure}{\small {\textbf{Approach Overview -- } Representative frames of $10$ shots from $2$ different scenes of the movie Stuart Little are shown. The story-arch of each scene is distinguishable and semantically coherent. We consider similar nearby shots (\textit{e.g.} $5$ and $3$) as augmented versions of each other. This augmentation approach is able to capitalize on the underlying film-production process and can encode the scene-structure better than the existing augmentation methods. Given a current shot (query) we find a similar shot (key) within its neighborhood and: (a) maximize the similarity between the query and the key, and (b) minimize the similarity of the query with randomly selected shots.}}\vspace{-0.3cm}
	\label{fig:teaser_fig}
}

\maketitle

\begin{abstract}

\vspace{-0.15cm}\noindent Scenes play a crucial role in breaking the storyline of movies and TV episodes into semantically cohesive parts. However, given their complex temporal structure, finding scene boundaries can be a challenging task requiring large amounts of labeled training data. To address this challenge, we present a self-supervised shot contrastive learning approach (ShotCoL) to learn a shot representation that maximizes the similarity between nearby shots compared to randomly selected shots. We show how to apply our learned shot representation for the task of scene boundary detection to offer state-of-the-art performance on the MovieNet~\cite{rao2020local} dataset while requiring only $\sim$$25\%$ of the training labels, using $9$$\times$ fewer model parameters and offering $7$$\times$ faster runtime. To assess the effectiveness of ShotCoL on novel applications of scene boundary detection, we take on the problem of finding timestamps in movies and TV episodes where video-ads can be inserted while offering a minimally disruptive viewing experience. To this end, we collected a new dataset called AdCuepoints with $3,975$ movies and TV episodes, $2.2$ million shots and $19,119$ minimally disruptive ad cue-point labels. We present a thorough empirical analysis on this dataset demonstrating the effectiveness of ShotCoL for ad cue-points detection.

%which consider similar nearby shots as augmented versions of each other and tries to maximize their similarity while minimizing it with randomly selected shots. This augmentation approach is able to exploit the underlying film-production process and can encode the local scene-structure in an effective manner. 
	
\end{abstract}

\vspace{-0.6cm}
\section{Introduction}

\noindent In filmmaking and video production, shots and scenes play a crucial role in effectively communicating a storyline by dividing it into easily interpretable parts. A \textbf{shot} is defined as a series of frames captured from the same camera over an uninterrupted period of time~\cite{sklar_film}, while a \textbf{scene} is defined as a series of shots depicting a semantically cohesive part of a story~\cite{katz_film} (see Figure~\ref{fig:teaser_fig} for an illustration). Localizing shots and scenes is an important step towards building semantic understanding of movies and TV episodes, and offers a broad range of applications including preview generation for browsing and discovery, content-driven video search, and minimally disruptive video-ads insertion.

Unlike shots which can be accurately localized using low-level visual cues~\cite{sidiropoulos2011temporal}~\cite{cotsaces2006video}, scenes in movies and TV episodes tend to have complex temporal structure of their constituent shots and therefore pose a significantly more difficult challenge for their accurate localization. Existing unsupervised approaches for scene boundary detection~\cite{chasanis2008scene} ~\cite{rasheed2003scene}~\cite{baraldi2015deep} do not offer competitive levels of accuracy, while supervised approaches~\cite{rao2020local} require large amounts of labeled training data and therefore do not scale well. Recently, several self-supervised learning approaches have been applied to learn generalized visual representations for images~\cite{jing2020self}~\cite{bachman2019learning}~\cite{henaff2019data}~\cite{hjelm2018learning}~\cite{oord2018representation}~\cite{wu2018unsupervised}~\cite{zhuang2019local}~\cite{tian2019contrastive} and short video clips~\cite{qian2020spatiotemporal}~\cite{han2020self}~\cite{wang2020enhancing}~\cite{tao2020self}, however it has been mostly unclear how to extend these approaches to long-form videos. This is primarily because the relatively simple data augmentation schemes used by previous self-supervised methods cannot encode the complex temporal scene-structure often found in long-form movies and TV-episodes.

To address this challenge, we propose a novel shot contrastive learning approach (ShotCoL) that naturally makes use of the underlying production process of long-form videos where directors and editors carefully arrange different shots and scenes to communicate the story in a smooth and believable manner. This underlying process gives rise to a simple yet effective invariance, \textit{i.e.}, nearby shots tend to have the same set of actors enacting a semantically cohesive story-arch, and are therefore in expectation more similar to each other than a set of randomly selected shots. This invariance enables us to consider nearby shots as augmented versions of each other where the augmentation function can implicitly capture the local scene-structure significantly better than the previously used augmentation schemes. Specifically, given a shot, we try to: (a) maximize its similarity with its most similar neighboring shot, and (b) minimize its similarity with a set of randomly selected shots (see Figure~\ref{fig:teaser_fig} for an illustration). 

We show how to use our learned shot representation for the task of scene boundary detection to achieve state-of-the-art results on  MovieNet dataset~\cite{rao2020local} while requiring only $\sim$$25\%$ of the training labels, using $9$$\times$ fewer model parameters, and offering $7$$\times$ faster runtime. Besides these performance benefits, our single-model based approach is significantly easier to maintain in a production setting compared to previous approaches that make use of multiple models~\cite{rao2020local}.

As a practical application of scene boundary detection, we explore the problem of finding timestamps in movies and TV episodes for minimally disruptive video-ads insertion. To this end, we present a new dataset called AdCuepoints with $3,975$ movies and TV episodes, $2.2$ million shots, and $19,119$ manually labeled minimally disruptive ad cue-points. We present a thorough empirical analysis on this dataset demonstrating the generalizability of ShotCoL on the task of ad cue-points detection.

\section{Related Work}

\vspace{0.1cm}\noindent \textbf{Self-Supervised Representation Learning:} 
Self supervised learning (SSL) is a class of algorithms that attempts to learn data representations using unlabeled data by solving a surrogate (or \textit{pretext}) task using supervised learning. Here the supervision signal for training can be automatically created~\cite{jing2020self} without requiring labeled data. Some of the previous SSL approaches have used the pretext task of reconstructing artificially corrupted inputs~\cite{vincent2008extracting}~\cite{pathak2016context}~\cite{zhang2016colorful}, while others have tried classifying inputs into a set of pre-defined categories with pseudo-labels~\cite{doersch2015unsupervised}~\cite{dosovitskiy2014discriminative}~\cite{pathak2017learning}.

\vspace{0.1cm}\noindent \textbf{Contrastive Learning:}
As an important subset of SSL methods, contrastive learning algorithms attempt to learn data representations by contrasting similar data against dissimilar data while using contrastive loss functions~\cite{le2020contrastive}. Contrastive learning has shown promise for multiple recognition based tasks for images~\cite{henaff2019data}~\cite{jing2020self}~\cite{chen2020simple}. Recently, with a queue-based mechanism that enables the use of large and consistent dictionaries in a contrastive learning setting, the momentum contrastive approach \cite{he2019moco}~\cite{chen2020mocov2} has demonstrated significant accuracy improvement compared to the earlier approaches. Recent works on using contrastive learning for video analysis~\cite{qian2020spatiotemporal}~\cite{han2020memory}\cite{Tschannen_2020_CVPR}\cite{Jayaraman_2016_CVPR} primarily focus on short-form videos where relatively simple data augmentation approaches have been applied to learn the pretext task. In contrast, our work focuses on long-form movies and TV episodes where we learn shot representations by incorporating a data augmentation mechanism that can exploit the underlying filmmaking process and therefore can encode the local scene-structure more effectively.

\vspace{0.1cm}\noindent \textbf{Scene Boundary Detection:}
Scene boundary detection is the problem of identifying the locations in videos where different scenes begin and end. Earlier methods for scene boundary detection such as~\cite{rui1998exploring}, adopt an unsupervised-learning approach that clusters the neighboring shots into scenes using spatiotemporal video features. Similar to~\cite{rui1998exploring}, the work in~\cite{rasheed2003scene} clusters shots based on their color similarity to identify potential boundaries, followed by a shot merging algorithm to avoid over-segmentation. More recently, supervised learning approaches~\cite{rotman2017optimal}~\cite{baraldi2015deep}~\cite{protasov2018using}~\cite{rao2020local} have been proposed to learn scene boundary detection using human-annotated labels. While these approaches offer better accuracy compared to earlier unsupervised approaches, they require large amounts of labeled training data and are therefore difficult to scale.

Multiple datasets have been used to evaluate scene boundary detection approaches. For instance, the OVSD dataset \cite{rotman2017optimal} includes $21$ videos with scene labels and scene boundaries. Similarly, the BBC planet earth dataset~\cite{baraldi2015deep} consists of $11$ documentaries labeled with scene boundaries. Recently, the MovieNet dataset~\cite{huang2020movienet} has taken a major step in this direction and published $1,100$ movies where $318$ of them are annotated with scene boundaries. Building on this effort to scale up the evaluation for scene boundary detection and its applications, we present empirical results on a new dataset called AdCuepoints with $3,975$ movies and TV episodes, $2.2$ million shots, and $19,119$ manual labels. 

%MovieNet dataset provides a large-scale foundation for video understanding research, and the state-of-the-art results on scene boundary detection is reported for the method based on deep neural networks with multi-modal inputs~\cite{Rao_2020_CVPR}.

\section{Method}
\label{sec:method}
\begin{figure*}[!htb]
	\centering
	\includegraphics[width=1.0\textwidth]{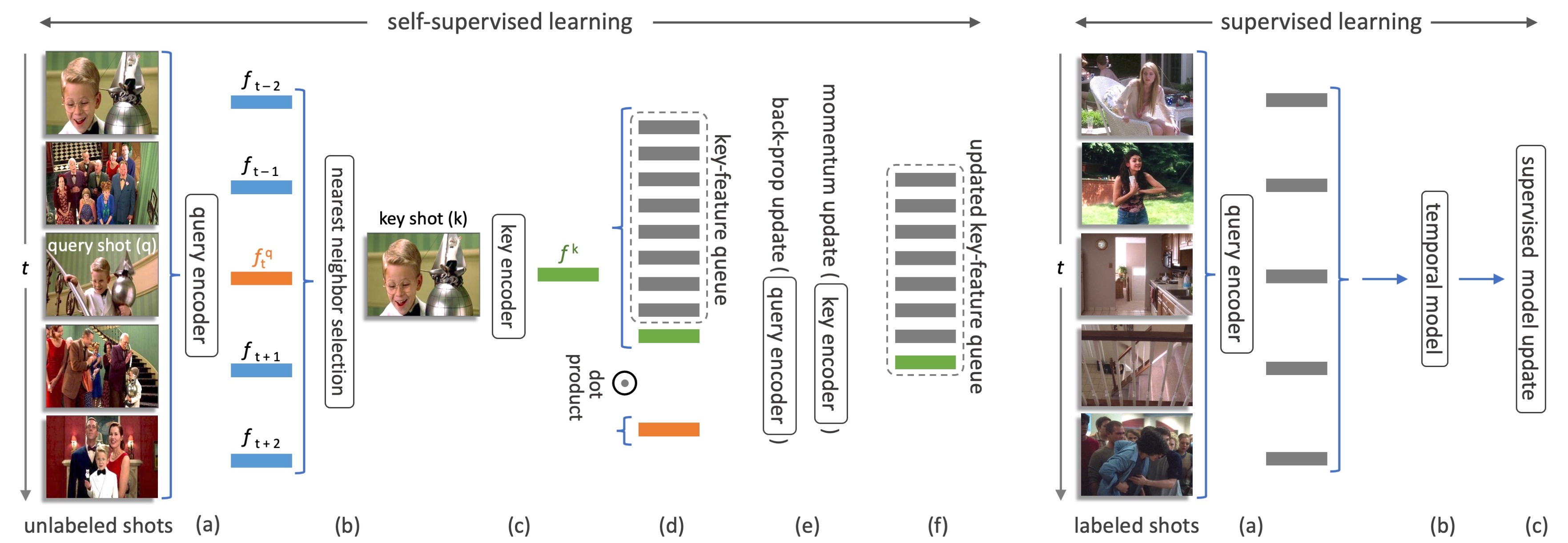}
	\caption{\textbf{Self-Supervised Learning}: (a) Use unlabeled data to extract the \textbf{visual or audio} features of a given query shot and its neighboring shots. (b) Find the key shot which is most similar to the query shot within its neighborhood. (c) Pass the key shot through the key encoder. (d) Contrast the query shot feature with key shot feature and the set of already queued features. (e) Use a contrastive loss function to update the query encoder through back-propagation and use momentum update for the key encoder. (f) Insert the key shot feature into the key-feature queue. \textbf{Supervised Learning}: (a) Use labeled data to extract visual or audio features of all shots by using the query encoder trained during the self-supervised learning step. (b) Learn temporal information among the shots. (c) Update the network using supervised learning.}
	\label{pipe}
\end{figure*}

\noindent We first discuss our self-supervised approach for shot-level representation learning where we present the details of our encoder network and contrastive learning approach. We then discuss how we use our trained encoder in a supervised learning setting for the task of scene boundary detection. Our overall approach is illustrated in Figure~\ref{pipe}.

\subsection{Shot-Level Representation Learning}
\noindent Given a full-length input video, we first use standard shot detection techniques~\cite{sidiropoulos2011temporal} to divide it into its constituent set of shots. Our approach for shot representation learning has two main components: (a) encoder network for visual and audio modalities, and (b) momentum contrastive learning~\cite{he2019moco} to contrast the similarity of the embedded shots. We now present the details for these two components.

\vspace{-0.4cm}\subsubsection{Shot Encoder Network}
\noindent We use separate encoder networks to independently learn representations for the audio and visual modalities of the input shots. Although ShotCoL is amenable to using any encoder network, the particular encoders we used in this work incorporate simplifications that are particularly conducive to scene boundary detection. The details of our visual and audio encoder networks are provided below.

\vspace{0.1cm} \noindent \textbf{1- Visual Modality: }Since scene boundaries exclusively depend on inter-shot relationships, encoding intra-shot frame-dynamics is not as important to us. We therefore begin by constructing a $4$D tensor $\mathit{(w,h,c,k)}$ from each shot with uniformly sampled $\mathit{k}$ frames each with $\mathit{w}$ width, $\mathit{h}$ height and $\mathit{c}$ color channels. We then reshape this $4$D tensor into a $3$D tensor by combining the $\mathit{c}$ and $\mathit{k}$ dimensions together. This conversion offers two key advantages:

\vspace{0.1cm}\noindent \textbf{a. Usage of Standard Networks}: As multiple standard networks (\textit{e.g.} AlexNet~\cite{krizhevsky2012imagenet}, VGG~\cite{simonyan2014very}, ResNet~\cite{ren2015faster}, \textit{etc.}) support $2$D images as input, by considering shots as $3$D tensors we are able to directly apply a wide set of standard image classification networks for our problem.

\vspace{0.1cm}\noindent \textbf{b. Resource Efficiency}: As we do not keep the time dimension explicitly after the first layer of our encoder network, we require less memory and compute resources compared to using temporal networks (\textit{e.g.} $3$D CNN~\cite{hara2018can}).

\vspace{0.1cm}\noindent Specifically, we use ResNet-$50$~\cite{he2016deep} as our encoder for the visual modality which produces a $2048$-dimensional feature vector to encode the visual signal for each shot.

\vspace{0.1cm} \noindent \textbf{2- Audio Modality: } To extract the audio embedding from each shot, we use a Wavegram-Logmel CNN~\cite{kong2020panns} which incorporates a $14$-layer CNN similar in architecture to the VGG~\cite{hershey2017cnn} network.
%and is pre-trained on the AudioSet~\cite{gemmeke2017audio} dataset.
We sample $10$-second mono audio samples at a rate of $32$ $\kilohertz$ from each shot. For shots that are less than $10$ seconds long, we equally zero-pad the left and right to form a $10$-second sample. For shots longer than $10$ seconds, we extract a $10$-second window from the center. These inputs are provided to the Wavegram-Logmel network~\cite{kong2020panns} to extract a $2048$-dimensional feature vector for each shot.
\label{pann}

\vspace{-0.4cm}\subsubsection{Shot Contrastive Learning}
\label{contrast_learning}
We apply contrastive learning~\cite{hadsell2006dimensionality} to obtain a shot representation that can effectively encode the local scene-structure and is therefore conducive for scene boundary detection. To this end, we propose to use a \textit{pretext}\footnote{We use the terms \textit{pretext}, \textit{query}, \textit{key} and \textit{pseudo-labels} as their standard usage in contrastive learning literature. See~\cite{he2019moco} for more information.} task that is able to exploit the underlying film-production process and encode the scene-structure better than the recent alternative video representations~\cite{qian2020spatiotemporal}~\cite{han2020self}~\cite{wang2020enhancing}~\cite{tao2020self}.

For a given \textit{query} shot, we first find the positive \textit{key} as its most similar shot within a neighborhood around the query, and then: (a) maximize the similarity between the query and the positive key, and (b) minimize the similarity of the query with a set of randomly selected shots (\textit{i.e.} negative keys). For this pretext task no human annotated labels are used. Instead, training is entirely based on the \textit{pseudo-labels} created when the pairs of query and key are formed.

\vspace{0.1cm}\noindent \textbf{a. Similarity and Neighborhood}:
More concretely, for a query at time $t$ denoted as $q_t$, we find its positive key $k_{0}$ as the most similar shot in a neighborhood consisting of $2$$\times$$\textrm{m}$ shots centered at $q_t$. This similarity is calculated based on the embeddings of the query encoder $f$$(\cdot| \theta_q)$:
\vspace{-0.1cm}
\begin{gather}
\label{select_k}
k_{0} = \arg\max_{x \in X_t} f(q_t | \theta_q) \cdot  f(x | \theta_q)\\
\vspace{-0.1cm}
X_t = [q_{t-m}, ..., q_{t-2}, q_{t-1}, q_{t+1}, q_{t+2}, ..., q_{t+m}]
\end{gather}
Along with $\textrm{K}$ negative keys $\textrm{S}_\textrm{K}$, the $\textrm{K}$+1 shots ($k_{0}$ $\cup$ $\textrm{S}_\textrm{K}$) are encoded by a key encoder to form a ($\textrm{K}$+1)-class classification task, where $q$ needs to be classified to class $k_{0}$.

\begin{figure}[!tb]
	\centering
	\includegraphics[width=0.48\textwidth]{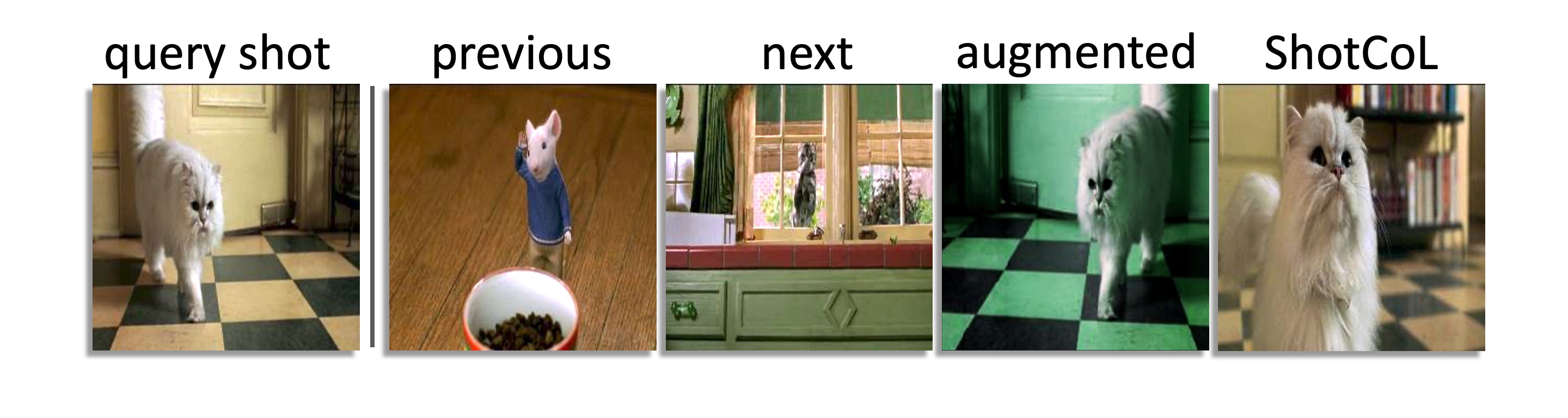}
	\vspace{-0.5cm}
	\caption{\small{Different ways to select positive key given a query shot.}}
	\label{augmentation}
	\vspace{-0.4cm}
\end{figure}

%\vspace{0.1cm}\noindent \textbf{b. Dictionary Look-Up}:
Our pretext task can be considered as training an encoder for a dictionary look-up task~\cite{he2019moco}, where given a query, the corresponding key should be matched. In our case, given an input query shot $q$, the goal is to find its positive key shot $k_{0}$ in a set of shots \{$k_{0}$, $k_{1}$, $k_{2}$, …, $k_{\textrm{K}}$\}. By defining the similarity as a dot product, we use the contrastive loss function InfoNCE \cite{oord2018representation}:

\begin{equation}
\label{loss}
\mathcal{L}_{q} = -\textrm{log} \frac{\textrm{exp}(f(q | \theta_q) \cdot g(k_0|\theta_k)/ \tau)}{\sum\limits_{i=0}^{\textrm{K}}\textrm{exp}(f(q | \theta_q) \cdot g(k_i|\theta_k) / \tau)}
\end{equation}
where $g$$(\cdot | \theta_k)$ is the key encoder with the parameter $\theta_k$. Here $k_{0}$ is the positive key shot, and $k_{1}$, $k_{2}$, …, $k_{\textrm{K}}$ are negative key shots. Also, $\tau$ is the temperature term~\cite{wu2018unsupervised} such that when $\tau=1$, Equation~\ref{loss} becomes standard log-loss function with softmax activation for multi-class classification.

The intuition behind our method of positive key selection is illustrated in Figure~\ref{augmentation}, where given a query shot, different ways to select its positive key are shown. Notice that using image-focused augmentation schemes (col. $4$) as done in \textit{e.g.}~\cite{he2019moco} does not incorporate any information about scene-structure. Similarly, choosing a shot adjacent to the query shot (col. $2$ and $3$) as the key can result in a too large and unrelated appearance difference between the query and key. Instead, selecting a similar nearby shot as the positive key provides useful information related to the scene-structure and therefore facilitates learning a useful shot representation. Results showing the ability of our shot representation to encode scene-structure are provided in $\S$~\ref{ss:rep_effectiveness}.

\vspace{0.1cm}\noindent \textbf{b. Momentum Contrast}:
Although large dictionaries tend to lead to more accurate representations, they also incur additional computational cost. To address this challenge,~\cite{he2019moco} recently proposed a queue-based solution to enable large-size dictionary training. Along similar lines, we save the embedded keys in a fixed-sized queue as negative keys. When a new mini-batch of keys come in, it is enqueued, and the oldest batch in the queue is dequeued. This allows the computed keys in the dictionary to be re-used across mini-batches.

To ensure consistency of keys when the key encoder evolves across mini-batch updates, a momentum update scheme~\cite{he2019moco} is used for the key encoder, with the following update equation:
\begin{equation}
\theta_k \leftarrow \alpha \cdot \theta_k + (1-\alpha) \cdot \theta_q
\end{equation}
\noindent where $\alpha$ is the momentum coefficient. As only $\theta_q$ is updated during back-propagation, $\theta_k$ can be considered as a moving average of $\theta_q$ across back-propagation steps.

\subsection{Supervised Learning}
\label{method_sup}
%As defined in previous work, scene is the fundamental unit for semantic understanding of movies and episodes \cite{rasheed2003scene, chasanis2008scene, han2011video, tapaswi2014storygraphs, rao2020local}.
\noindent Recall that scenes are composed of a series of contiguous shots. Therefore, we formulate the problem of scene boundary detection as a binary classification problem of determining if a shot boundary is also a scene boundary or not.

To this end, after dividing a full length video into its constituent shots using low-level visual cues~\cite{sidiropoulos2011temporal}, for each shot boundary we consider its $2\times\textrm{N}$ neighboring shots ($\textrm{N}$ shots before and $\textrm{N}$ shots after the shot boundary) as a data-point to perform scene boundary detection.

For each data-point, we use the query encoder trained by contrastive learning to extract shot-level visual or audio features independently. We then concatenate the feature vectors of the $2\times\textrm{N}$ shots into a single vector, which is then provided as an input to a multi-layer perceptron (MLP) classifier\footnote{Note that other classifiers besides MLP can also be used here. See $\S$~\ref{adcuepoints_temporal_results} for comparative results of using different temporal models.}. The MLP consists of three fully-connected (FC) layers where the final FC layer is followed by softmax for normalizing the logits from FC layers into class probabilities of the positive and negative classes. Unless otherwise mentioned, the weights of the trained encoder are kept fixed during this step, and only MLP weights are learned.

%Also note that other models besides MLP can also be used at this step (see $\S$blah for comparative results). Furthermore, 

During inference, for each shot boundary, we form the $2\times\textrm{N}$-shot sample, extract shot feature vectors and pass the concatenated feature to our trained MLP to predict if the shot boundary is a scene boundary or not.
\section{Experiments}
\label{sec:experiments}

\noindent We first present results to distill the effectiveness of our learned shot representation in terms of its ability to encode the local scene-structure, and then use detailed comparative results to show its competence for the task of scene boundary detection. Finally, we demonstrate the results of ShotCoL for a novel application of scene boundary detection, \textit{i.e.} finding minimally disruptive ad cue-points.

\subsection{Effectiveness of Learned Shot Representation}
\label{ss:rep_effectiveness}
\begin{figure}[!t]
	\centering
	\includegraphics[width=0.4125\textwidth]{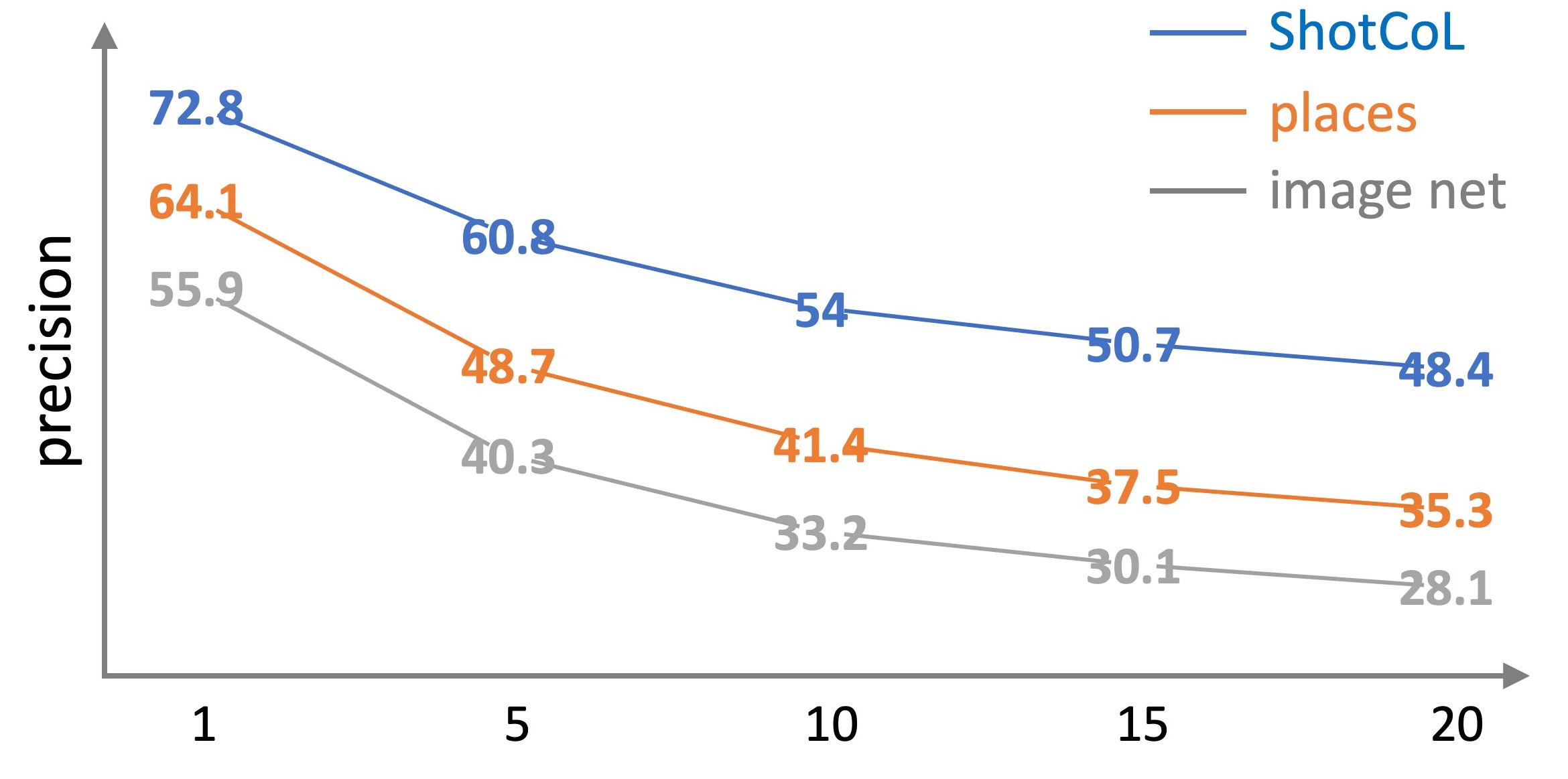}
	\vspace{-0.2cm}
	\caption{\small{Comparison of shot retrieval precision (y-axis) using the test split of the MovieNet dataset~\cite{huang2020movienet} with different number of nearest neighbors (x-axis) and shot representations.}}\vspace{-0.3cm}
	\label{NN}
\end{figure}

\noindent Intuitively, if a shot representation is able to project shots from the same scenes to be close to each other, it should be useful for scene boundary detection. To test how well our learned shot representation is able to do this, given a query shot from a movie, we retrieve its $k$ nearest neighbor shots from the same movie. Retrieved shots belonging to the same scene as the query shot are counted as true positives, while those from other scenes as false positives. We use the test split of MovieNet~\cite{huang2020movienet}, and compare our learned shot representation (see $\S$~\ref{scene_bd} for details) with Places~\cite{zhou2017places} and ImageNet~\cite{deng2009imagenet} features computed using ResNet-50~\cite{he2016deep}.

Results in Figure~\ref{NN} show that our learned shot representation significantly outperforms other representations for a wide range of neighborhood sizes, demonstrating its ability to encode the local scene-structure more effectively.

\begin{figure}[!t]
	\centering
	\includegraphics[width=0.48\textwidth]{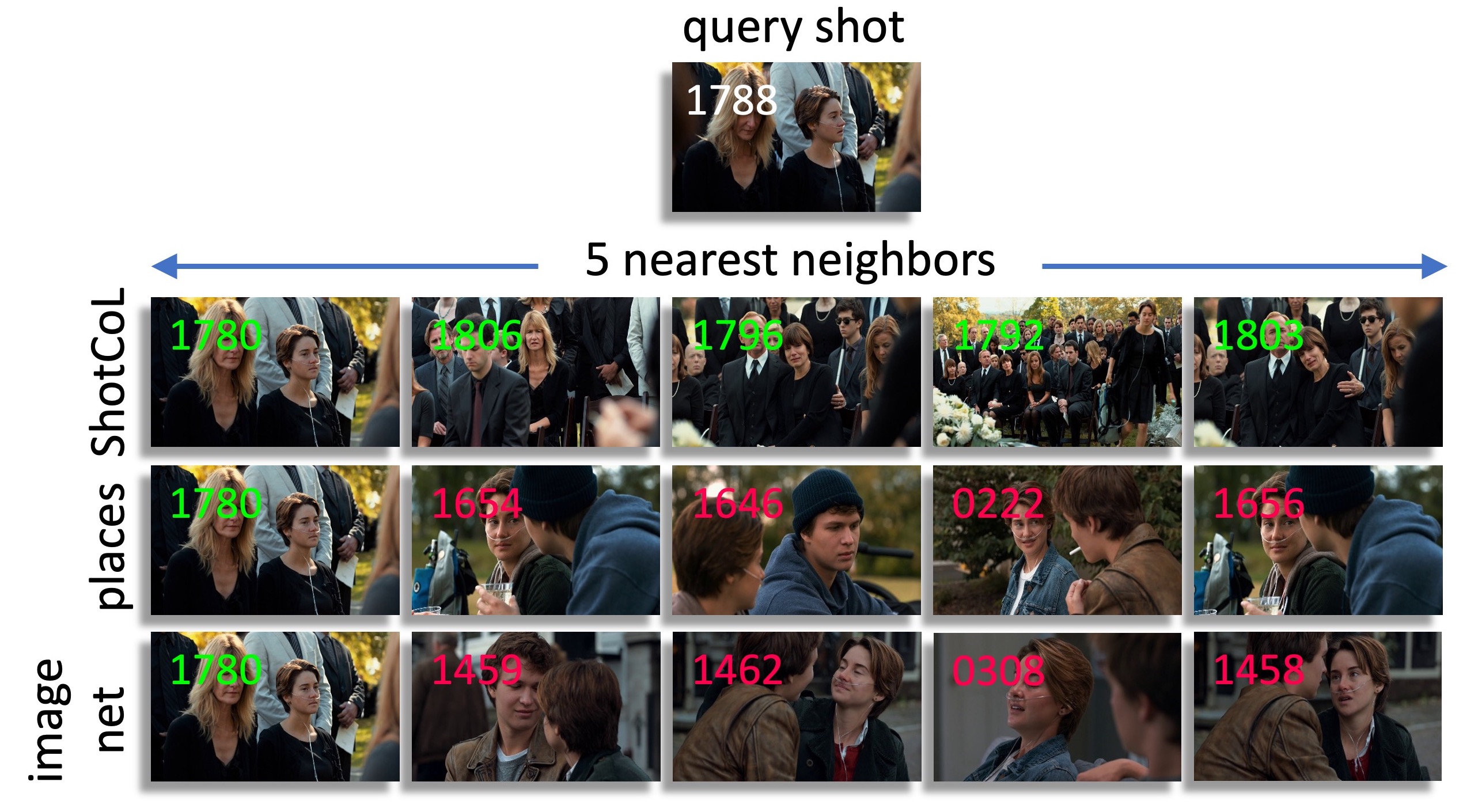}
	\vspace{-0.7cm}
	\caption{\small{Five nearest neighbor shots for a query shot using different shot representations are shown. Shot indices are displayed at top-left corners where green indicates shot from same scene as query while red indicates shot from a different scene.}}\vspace{-0.5cm}
	\label{examples}
\end{figure}

Figure~\ref{examples} provides an example qualitative result where $5$ nearest neighbor shots for a query shot using different shot representations are shown. While results retrieved using Places~\cite{zhou2017places} and ImageNet~\cite{deng2009imagenet} features are visually quite similar to the query shot, almost none of them are from the query shot's scene. In contrast, results from ShotCoL representation are all from the same scene even though the appearances of the retrieved shots do not exactly match query shot. This shows that our learned shot representation is able to effectively encode the local scene-structure.

\subsection{Scene Boundary Detection}
\label{scene_bd}
\noindent We now present comparative performance of various models for scene boundary detection using MovieNet data~\cite{huang2020movienet}.

\noindent \textbf{a. Evaluation Metrics:}
We use the commonly used metrics to evaluate the considered methods~\cite{rao2020local}, \textit{i.e.} Average Precision (AP), Recall and  Recall@$3$s, where Recall@$3$s calculates the percentage of the ground truth scene boundaries which are within $3$ seconds of the predicted ones.

\vspace{0.1cm}\noindent \textbf{b. Dataset:}
Our comparative analysis for scene boundary detection uses the MovieNet dataset~\cite{huang2020movienet} which has $1,100$ movies out of which $318$ have scene boundary annotations. The train, validation and test splits for these $318$ movies are already provided by authors of MovieNet~\cite{huang2020movienet} with $190$, $64$ and $64$ movies respectively. The scene boundaries are annotated at shot level and three key frames are provided for each shot. Following~\cite{rao2020local}, we report the metrics on only the test set for all of our experiments unless otherwise specified.

\vspace{0.1cm}\noindent \textbf{c. Implementation Details:}
We use all $1,100$ movies with $\sim$$1.59$ million shots in MovieNet~\cite{huang2020movienet} to learn our shot representation, and $190$ movies with scene boundary annotations to train our MLP classifier. All weights in the encoder and MLP are randomly initialized.
For contrastive learning settings, as 80\% of all scenes in MovieNet are $16$ shots or less, we fix the neighborhood size for positive key selection to $8$ shots.
%For our contrastive learning setting, the neighborhood size $m$ is fixed to $8$ for positive key selection based on the distribution of scene length in the dataset, and 
Other hyper-parameters are similar to MoCo~\cite{he2019moco}, \textit{i.e.}, $65,536$ queue size, $0.999$ MoCo momentum, and $0.07$ softmax temperature.
The initial set of positive keys is selected based on the space of ImageNet (details in Supplementary Material).
We use a three-layer MLP classifier (number-of-shots-used$\times$$2048$-$4096$-$1024$-$2$), and use dropout after each of the first two FC layers.
%$200$ epochs, $256$ batch size, $0.03$ learning rate, $0.9$ momentum, 

\vspace{-0.2cm}\subsubsection{Ablation Study}
Focusing on visual modality, we evaluate ShotCoL on the validation set of MovieNet for: (a) different number of shots, and (b) different number of key frames used per shot. As shown in Table \ref{ablation}, using $2$ shots in ShotCoL does not perform well signifying that the context within $2$ shots is not enough for classifying scene boundaries accurately. The features using $4$ shots achieve the highest AP, however the AP decreases when more shots are included. This is because as the context becomes larger, there is a higher chance of having multiple scene boundaries in each sample which makes the task more challenging for the model. In terms of the number of keyframes, the shot representation learned using $3$ keyframes performs better than the one using only $1$ keyframe. This indicates that the subtle temporal relationship within each shot can be beneficial for distinguishing different scenes.

\begin{table}[!t]
	\begin{center}
		\begin{tabu}{c!{\vrule width 1.25pt}ccccc}
			\hline
			 \# of &	\multicolumn{5}{c}{\# of shots}\\\cline{2-6}
			keyframes& $2$ & $4$ & $6$ & $8$ & $10$\\\tabucline[1.25pt]{-}
			$1$ &48.66	&55.24	&54.89	&53.89	&52.94\\\hline
			$3$ &48.95	&\textbf{56.13}	&55.73	&54.01	&53.07\\\hline
		\end{tabu}
	\end{center}
	\vspace{-0.5cm}\caption{\small{AP results for ablation study in MovieNet data~\cite{huang2020movienet}}.}\vspace{-0.3cm}
	\label{ablation}
\end{table}

Based on this ablation study, for all our experiments we use $3$ frames per shot. For all of our scene boundary detection experiments we use a context of $4$ shots (two to the left and two to the right) around each shot transition point to form a positive or negative sample based on its label.

\begin{table*}[!htb]
	\begin{center}
		\begin{tabu}{cccccccc}
			\hline
			&\multirow{2}{*}{Models}	&\multirow{2}{*}{Modalities} & Est. \# of Encoder	& Est. inference 		&	\multirow{2}{*}{AP}  & {Recall}& {Recall@3s}\\
			&&& Parameters&time/batch&&(0.5 thr.)&(0.5 thr.)\\\tabucline[1.25pt]{-}
			\multicolumn{8}{c}{Without self-supervised pre-training}\\
			\hline
			1 &SCSA \cite{chasanis2008scene}&Visual&23 m&6.6s&14.7&54.9&58.0\\
			2&Story Graph \cite{tapaswi2014storygraphs}&Visual&23 m&6.6s&25.1&58.4&59.7\\
			3&Siamese \cite{baraldi2015deep}&Visual& 23 m&6.6s&28.1&60.1&61.2\\
			4&ImageNet \cite{deng2009imagenet}&Visual&23 m&2.64s&41.26&30.06&33.68\\
			5&Places \cite{zhou2017places} & Visual & 23 m & 2.64s & 43.23&59.34&64.62\\
			\hline
			\multirow{2}{*}{6}&\multirow{2}{*}{LGSS \cite{rao2020local}}&Visual Audio&\multirow{2}{*}{228 m}&\multirow{2}{*}{39.6s}&\multirow{2}{*}{47.1}&\multirow{2}{*}{73.6}&\multirow{2}{*}{79.8}\\
			&&Action Actor&&&&&\\\tabucline[1.25pt]{-}
			\multicolumn{8}{c}{With self-supervised pre-training} \\\hline
			\multirow{2}{*}{7}&SimCLR \cite{chen2020simple}  &\multirow{2}{*}{Visual}&\multirow{2}{*}{23 m}&\multirow{2}{*}{2.64s}&\multirow{2}{*}{41.65}&\multirow{2}{*}{75.01}&\multirow{2}{*}{80.42}\\
			&(img. aug.)&&&&&&\\\hline
			\multirow{2}{*}{8}&MoCo \cite{he2019moco} &\multirow{2}{*}{Visual}&\multirow{2}{*}{23 m}&\multirow{2}{*}{2.64s}&\multirow{2}{*}{42.51}&\multirow{2}{*}{71.53}&\multirow{2}{*}{77.11}\\
			&(img. aug.) &&&&&&\\\hline
			\multirow{2}{*}{9}&SimCLR \cite{chen2020simple} &\multirow{2}{*}{Visual}&\multirow{2}{*}{25 m}&\multirow{2}{*}{5.39s}&\multirow{2}{*}{50.45}&\multirow{2}{*}{81.31}&\multirow{2}{*}{\textbf{85.91}}\\
			&(shot similarity)  &&&&&&\\\hline
			\multirow{2}{*}{10}&ShotCoL &\multirow{2}{*}{Visual}&\multirow{2}{*}{25 m}&\multirow{2}{*}{5.39s}&52.83&81.59
			&85.44\\
			&(MovieScenes \cite{rao2020local}) &&&&$\pm$2.08&$\pm$1.82&$\pm$1.46\\\hline
			11&ShotCoL&Visual&25 m&5.39s&\textbf{53.37}&\textbf{81.33}&85.34\\\hline
			
		\end{tabu}
	\end{center}
	\vspace{-0.3cm}\caption{\small{\textbf{Comparative analysis for scene boundary detection -- } The compared methods are grouped in two, \textit{i.e.}: (a) ones that do not use self-supervised learning and (b) ones that use self-supervised learning followed by use of learned features in a supervised setting.}}
	\label{comparison_mn}
\end{table*}

\vspace{-0.2cm}\subsubsection{Comparative Empirical Analysis}

%\vspace{0.1cm}\noindent \textbf{a. Full Data-Set:}

%As baseline, we extract ResNet-$50$~\cite{he2016deep} features pre-trained on ImageNet~\cite{deng2009imagenet} to learn the MLP classifier. 
\noindent The detailed comparative results are given in~Table~\ref{comparison_mn}. LGSS~\cite{rao2020local} has been the state-of-the-art on the MovieNet data~\cite{huang2020movienet} reporting $47.1$ AP achieved by using four pre-trained models (two ResNet-$50$, one ResNet-$101$ and one VGG-m) on multiple modalities together with LSTM. We comfortably outperform LGSS~\cite{rao2020local} (relative margin of $13.3\%$) using a single network on visual modality only. Moreover, ShotCoL offers $9$$\times$ fewer model parameters and $7$$\times$ faster runtime compared to LGSS~\cite{rao2020local}. 

Recall that results in~\cite{rao2020local} were reported using $150$ titles from MovieNet~\cite{huang2020movienet} with $100$, $20$ and $30$ titles for training, validation and testing respectively. Therefore, we  also provide results on the $150$ titles subset of MovieNet~\cite{huang2020movienet} (called MovieScenes \cite{rao2020local}). As the exact data-splits are not provided by~\cite{rao2020local}, we do a $10$-fold cross-validation and report the mean and standard deviation, showing $12.1\%$ relative performance gain over ~\cite{rao2020local} in expectation.

To compare our shot contrastive learning with previous self-supervised methods, we focus on two recently proposed methods outlined in~\cite{he2019moco} and~\cite{chen2020simple}. For each of these approaches, we consider two types of data augmentation strategies: (a) traditional image augmentation schemes (as used in~\cite{he2019moco} and~\cite{chen2020simple}), and (b) our proposed shot augmentation scheme. Results in Table~\ref{comparison_mn} show that using image-focused augmentation schemes only marginally improves the performance over the ImageNet baseline. In contrast, using our proposed shot augmentation scheme with either~\cite{he2019moco} or~\cite{chen2020simple} results in significant improvements.

%For comparison with approach in~\cite{he2019moco}, we contrast shots with their self-augmented versions in a momentum contrastive setting~\cite{he2019moco}. For comparison with~\cite{chen2020simple}, we learn shot similarly with a simpler contrastive learning approach where unlike~\cite{he2019moco} we have a single encoder with an end-to-end update. We can see that with Moco-based image augmentation approach, the improvement is limited compared to ImageNet feature, with shot similarity+DrLIM, the performance can be improved a lot mainly due to the usage of the proposed shot similarity invariance.

\begin{figure}[!t]
	\centering
	\includegraphics[width=0.425\textwidth]{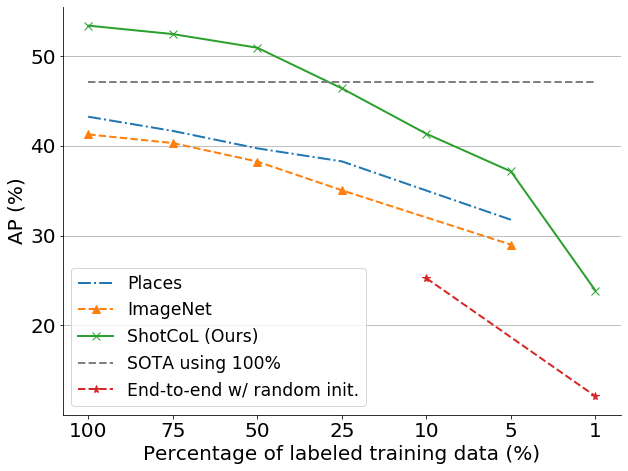}
	\vspace{-0.1cm}
	\caption{\small{Results on different label-amounts for MovieNet~\cite{huang2020movienet} data. Dashed Gray line is for LGSS~\cite{rao2020local} with 100\% labels.}}\vspace{-0.5cm}
	\label{curve_mn}
\end{figure}

\vspace{0.1cm}\noindent \textbf{Limited Amount of Labeled Training Data:}
The comparative performance of using our learned shot representation in limited labeled settings is given in Figure \ref{curve_mn}. Our learned feature is able to achieve $47.1$ test AP (results reported by LGSS~\cite{rao2020local}) while using only $\sim$$25\%$ of training labels.

Moreover, we compare the performance of ShotCoL with an end-to-end learning based setting with limited labeled data following the protocols in \cite{chen2020simple}. As shown in Figure \ref{curve_mn}, learning an end-to-end model with random initialization and limited training labels is challenging. Instead, ShotCoL is able to achieve significantly better performance using limited number of training labels.

\subsection{Application -- Ad Cue-Points Detection}
\noindent To assess the effectiveness of ShotCoL on novel applications of scene boundary detection, we take on the problem of finding timestamps in movies and TV episodes where video-ads can be inserted while being minimally disruptive. Such timestamps are referred to as \textit{ad cue-points}, and are required to follow multiple constraints. First, cue-points can only occur when the context of the storyline clearly and unambiguously changes. This is different from scene boundaries observed in other datasets such as MovieNet~\cite{huang2020movienet}, where the before and after parts of a scene can be contextually closely related. Second, cue-points cannot have dialogical activity in their immediate neighborhood. Third, all cue-points must be a certain duration apart and their total number needs to be within a certain limit which is a function of the video length. These constraints make ad cue-points detection a special case of scene boundary detection.

\begin{figure*}[!t]
	\centering
	\includegraphics[width=1.0\textwidth]{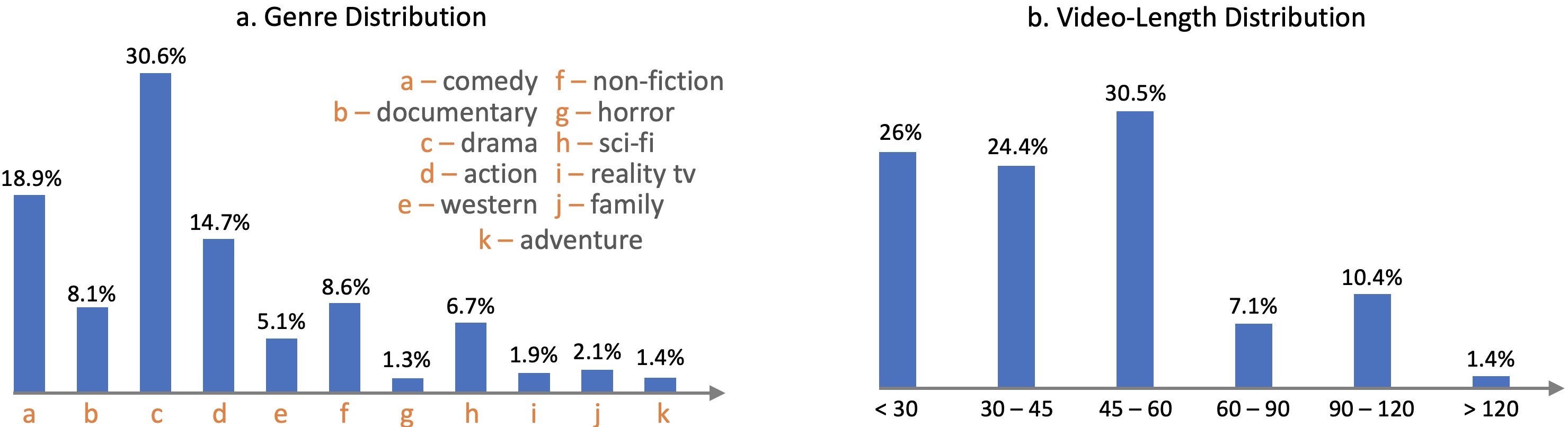}
	\vspace{-0.3cm}\caption{\small{\textbf{a --} Distribution of video genres in AdCuepoints dataset. \textbf{b --} Distribution of video length in AdCuepoints dataset.}}\vspace{-0.3cm}
	\label{durations}
\end{figure*}

%\begin{figure}[!t]
%	\centering
%	\includegraphics[width=0.4\textwidth]{genre_distribution_1.jpg}
%	\vspace{-0.2cm}\caption{\small{Distribution of video-genres in AdCuepoints data-set.}}\vspace{-0.2cm}
%	\label{genre}
%\end{figure}

%\begin{figure}[!t]
%	\centering
%	\includegraphics[width=0.4\textwidth]{duration_distribution.jpg}
%	\vspace{-0.2cm}\caption{\small{Distribution of video-length in AdCuepoints data-set.}}\vspace{-0.3cm}
%	\label{duration}
%\end{figure}

%, and present a comprehensive empirical analysis of multiple approaches for this problem.

%The goal of Ad cuepoint detection is to detect the places in the video where the Ads can be inserted with the minimum watching disruption for video viewers. The Ad cuepoints are subset of scene boundaries as the Ad cuepoint is definitely a scene boundary but not all scene boundaries are Ad cuepoints. To facilitate the research of this direction, we propose the AdCuepoints dataset and provide the state-of-the-art results on it based on our model.

\vspace{-0.2cm}\subsubsection{AdCuepoints Dataset}
The AdCuepoints dataset contains $3,975$ movies and TV episodes, $2.2$ million shots, and $19,119$ manually labeled cue-points. Compared to the MovieNet dataset~\cite{huang2020movienet} which only contains movies, the AdCuepoints dataset also contains TV episodes which makes it more versatile and diverse from a content-based perspective. The video distribution of various genres present in AdCuepoints dataset is given in Figure \ref{durations}-a. The distribution of video lengths in the AdCuepoints dataset is provided in Figure~\ref{durations}-b.

We divide the $3,975$ full-length videos in the AdCuepoints dataset into their constituent shots by applying commonly used shot detection approaches~\cite{sidiropoulos2011temporal}. Recall that cue-points always lie at shot boundaries. We consider $k$ shots to the left and right of each cue-point to create a positive sample with $\pm k$ context. Negative samples are created around positive samples by taking a sliding-window traversal to the left and right of positive samples while incorporating a unit stride. We divide our dataset into training, validation and testing sets with $70$\%, $10$\%, $20$\% ratio respectively.

\subsubsection{Results}
\noindent \textbf{a. Visual Modality:} We learn our shot representation using the entire unlabeled AdCuepoints dataset, and then apply it along with other representations as inputs to MLP models that use cue-point labels for training. Table~\ref{ac_feats} shows that our shot representation performs significantly better than the alternatives. Here ImageNet~\cite{deng2009imagenet} features on $2$D-CNN \cite{he2016deep} and Kinetics~\cite{kay2017kinetics} features on $3$D-CNN \cite{hara2018can} provide baselines.

Note that even when using the shot similarity features self-trained on unlabeled MovieNet data~\cite{huang2020movienet}, the results obtained on AdCuepoints test data are significantly better than baseline. Similar trend can be observed on the cross-dataset setting of training our shot representation on unlabeled AdCuepoints data and testing on the MovieNet data~\cite{huang2020movienet}, where we achieve 48.40\% AP.  These results demonstrate that our learned shot representation is able to generalize well in a cross-dataset setting.

\begin{table}[!t]
	\begin{center}
		\begin{tabu}{cc|c|c}
			\hline
			\multicolumn{3}{c|}{Visual Feature}	&  \multirow{2}{*}{AP} \\\cline{1-3}
			&Pre-training data & Labeled data &\\\tabucline[1.25pt]{-}
			1 &ImageNet \cite{deng2009imagenet} &AdCuepoints	&45.90\\
			2&Kinetics \cite{kay2017kinetics} &AdCuepoints	&46.33\\
			3&AdCuepoints &AdCuepoints&\textbf{53.98}\\
			4&MovieNet &AdCuepoints&51.32\\
			5&AdCuepoints &MovieNet&48.40\\\hline
		\end{tabu}
	\end{center}
	\vspace{-0.4cm}\caption{\small{Performance of using different visual features on AdCuepoints dataset and cross-dataset results.}}\vspace{-0.1cm}
	\label{ac_feats}
\end{table}

\begin{table}[!t]
	\begin{center}
		\begin{tabu}{c|ccccc}
			\hline
			Audio  &	\multicolumn{5}{c}{\# of shots}\\\cline{2-6}
			Feature & $2$ & $4$ & $6$ & $8$ & $10$\\\tabucline[1.25pt]{-}
			PANN \cite{kong2020panns}&43.56	&46.47	&47.17	&46.97	&47.40\\
			ShotCoL &49.38	&52.56	&52.7	&53.45	&53.27\\\hline
		\end{tabu}
	\end{center}
	\vspace{-0.5cm}\caption{\small{Performance comparison of using pre-trained audio features~\cite{kong2020panns} with ShotCoL based audio feature.}}\vspace{-0.1cm}
	\label{adcuepoints_non_temporal_results}
\end{table}

\begin{table}[!t]
	\begin{center}
		\begin{tabu}{c|c|c|c}
			\hline
			&\multirow{2}{*}{MLP} &B-LSTM &Linformer \\
			&& \cite{huang2015bidirectional} + MLP&\cite{wang2020linformer} + MLP\\\tabucline[1.25pt]{-}
			\# of shots& 4 & 10 & 10\\\hline
			\# of parameters&71 m&197 m&190 m\\\hline
			AP & 57.65 & 59.02 &  59.95\\\hline
		\end{tabu}
	\end{center}
	\vspace{-0.5cm}\caption{\small{Comparison with different temporal models on the combined audio-visual feature.}}\vspace{-0.1cm}
	\label{adcuepoints_temporal_results}
\end{table}

\vspace{0.1cm} \noindent \textbf{b. Audio Modality:} Following the aforementioned procedure of our visual modality comparison, Table~\ref{adcuepoints_non_temporal_results} presents the results of using pre-trained audio features~\cite{kong2020panns} compared with audio features learned using ShotCoL on AdCuepoints dataset. Results using different number of shots are presented showing that ShotCoL is able to outperform existing approach by a sizable margin, demonstrating its effectiveness on audio modality.

%Table~\ref{adcuepoints_non_temporal_results} presents the results of using pre-trained audio features~\cite{kong2020panns} compared with audio features obtained using our contrastive learning approach on AdsCuepoints data-set using different number of shots. Our approach is able to out-perform existing approach by a sizable margin, demonstrating its effectiveness on audio modality.

\begin{figure}[!t]
	\centering
	\includegraphics[width=0.46\textwidth]{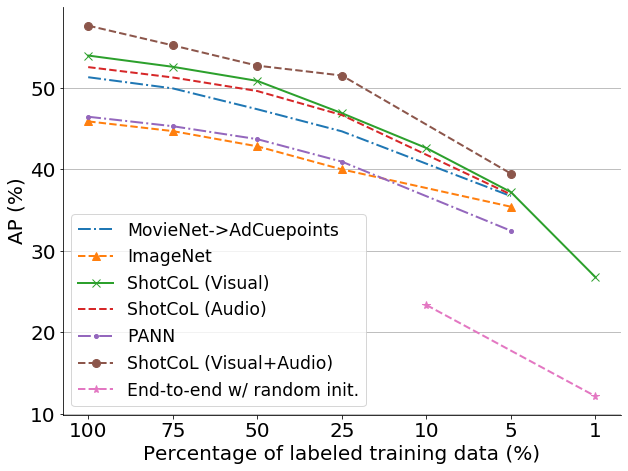}
	\vspace{-0.1cm}
	\caption{\small{Comparative performance when using different amounts of training data for AdCuepoints dataset.}}\vspace{-0.2cm}
	\label{curve_cp}
\end{figure}

\vspace{0.1cm} \noindent \textbf{c. Audio-Visual Fusion:} Table~ \ref{adcuepoints_temporal_results} shows how combining learned audio and visual features can help further improve the performance of ShotCoL. Column $1$ shows the result for concatenating our learned audio and visual shot representations and providing them as input to an MLP model. Moreover, columns $2$ and $3$ demonstrate that incorporating more sophisticated temporal models (\textit{i.e.} B-LSTM~\cite{huang2015bidirectional} and Linformer\cite{wang2020linformer}) can help fuse the audio and visual modalities more effectively than using simple feature concatenation. This shows that our shot representation can be used with a broad class of models downstream. 

%Table~ \ref{adcuepoints_temporal_results} in column $1$ shows the results of combining the learned shot-representation for audio and visual modalities by simple concatenation and as input to an MLP. On top of that, if we utilize some most recent temporal models like B-LSTM \cite{huang2015bidirectional} or \cite{wang2020linformer}, the performance can be further improved. Since the estimator part is not our major concern, and in fact it's orthogonal to our approach, we just briefly demonstrate the generality of our approach in this way.

\vspace{0.1cm} \noindent \textbf{d. Limited Amount of Labeled Data:}
The comparison of different shot representations when using limited amounts of labeled training data is provided in Figure~\ref{curve_cp}. It can be observed that ShotCoL is able to comfortably outperform all other considered methods on the AdCuepoints dataset. Moreover, we compare ShotCoL with an end-to-end learning setting as \cite{chen2020simple}, where we use only $10\%$ and $1\%$ of the labeled training data. It can be seen that using our learned features with limited labeled data is able to give significantly better performance compared to using end-to-end learning.

\section{Conclusions and Future Work}

\noindent We presented a self-supervised learning approach to learn a shot representation for long-form videos using unlabeled video data. Our approach is based on the key observation that nearby shots in movies and TV episodes tend to have the same set of actors enacting a cohesive story-arch, and are therefore in expectation more similar to each other than a set of randomly selected shots. We used this observation to consider nearby similar shots as augmented versions of each other and demonstrated that when used in a contrastive learning setting, this augmentation scheme can encode the scene-structure more effectively than existing augmentation schemes that are primarily geared towards images and short videos. We presented detailed comparative results to demonstrate the effectiveness of our learned shot representation for scene boundary detection. To test our approach on a novel application of scene boundary detection, we take on automatically finding ad cue-points in movies and TV episodes and use a newly collected large-scale data to show the competence of our method for this application.

Going forward, we will focus on improving the efficiency of contrastive video representation learning. We will also investigate the application of our shot representation to additional problems in video understanding.

%and demonstrate that this 
%This invariance leads us to our contrastive learning algorithm with the surrogate task of finding the most similar shot in a dictionary comprised of neighboring shots to the query and randomly selected shots.
%while distant shots are often dissimilar.
%In this work, we proposed a self-supervised approach to learn a contrastive shot-representation that makes use of the underlying film 
%this invariance in a self-supervised learning setting to consider nearby similar shots as augmented versions of each other in order to learn a contrastive shot-representation. We present a thorough empirical analysis to demonstrate the effectiveness of our learned shot-representation for the task of scene boundary detection. Moreover, to test our approach on a novel application of scene boundary detection, we take on the problem of automatically finding ad cue-points timestamps in movies and TV episodes and use a newly collected large-scale AdCuepoints to demonstrate the effectiveness of our approach for the ads cue-points detection problem.

{\small
\bibliographystyle{ieee_fullname}
\bibliography{ref}
}

\end{document}

\title{Supplementary Material:\\ Shot Contrastive Self-Supervised Learning for Scene Boundary Detection}

\author{Shixing Chen \quad Xiaohan Nie\thanks{Equal contribution.} \qquad David Fan\footnotemark[1] \qquad Dongqing Zhang \quad Vimal Bhat \quad Raffay Hamid\\
	Amazon Prime Video\\
	{\tt\small \{shixic, nxiaohan, fandavi, zdongqin, vimalb, raffay\}@amazon.com}
}

\maketitle

\noindent In this supplementary material, we: (a) present further details of our experimental settings for better reproducibility, (b) provide additional quantitative results for more comprehensive analysis, and (c) show additional qualitative results to provide a better understanding of our shot contrastive learning approach (ShotCoL).

\section{Experiment Details}

\noindent We used PyTorch~\cite{paszke2019pytorch} and Tesla V$100$ GPUs for all our experiments. For contrastive learning, we used $8$ GPUs with the  PyTorch module \textrm{DistributedDataParallel}. Below, we provide the network parameters, hyper-parameters and training details for the experiments on each dataset.

\subsection{MovieNet Dataset}

\noindent This section corresponds to $\S4.2.2$ in the main paper.

\vspace{-0.2cm}\subsubsection{Training Details}

\vspace{-0.1cm}\paragraph{a. Contrastive Learning:}
\label{contrast_vis}

\noindent Recall that we used ResNet-$50$ \cite{he2016deep} with the first layer modified to take $9$ input channels as our visual encoder. The weights of the query encoder $\theta_q$  were randomly initialized. The weights of the key encoder $\theta_k$ were initialized to the same values as $\theta_q$.
The query encoder was trained using SGD with a mini-batch size of $256$, momentum of $0.9$, and weight decay of $0.0001$. The initial learning rate of $0.03$ was dropped twice  each time by $10\times$.

\vspace{-0.2cm}\paragraph{b. Supervised Learning:}

For training the multi-layer perceptron (MLP) for the scene boundary detection task on the MovieNet \cite{huang2020movienet} dataset, we used dropout value of $0.9$, batch size of $1024$, maximum epoch number of $200$ and SGD with a fixed learning rate of $0.1$.

\subsubsection{Additional Results}

\begin{figure}[!htb]
	\centering
	\includegraphics[width=0.39\textwidth]{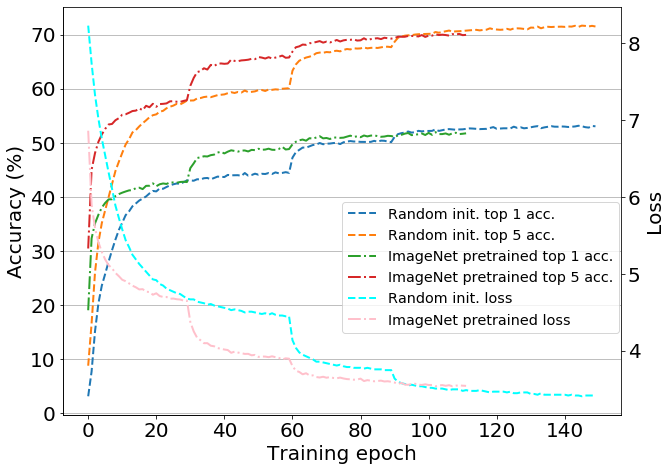}
	\vskip -0.1in
	\caption{Training curves of contrastive learning on MovieNet dataset.}
	\label{mn_tr_curve}
	\vskip -0.1in
\end{figure}

\begin{figure}[!htb]
	\centering
	\includegraphics[width=0.39\textwidth]{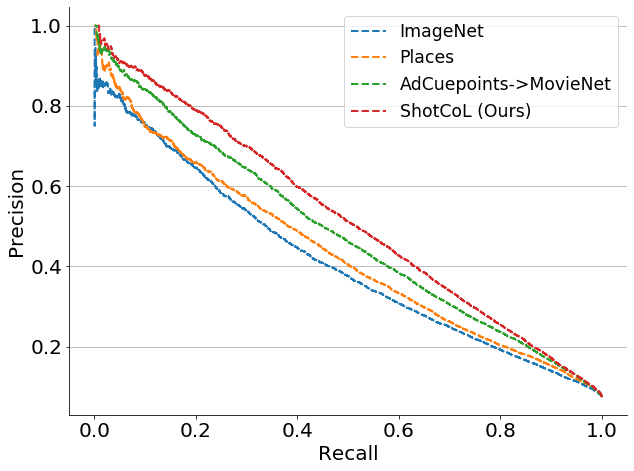}
	\vskip -0.1in
	\caption{PR-curves of test set in MovieNet dataset.}
	\label{mn_pr_curve}
	\vskip -0.15in
\end{figure}

\paragraph{a. Visual Modality:} We present the top $1$ accuracy, top $5$ accuracy and loss curves during contrastive training in Figure \ref{mn_tr_curve}.
There are two types of weight initialization methods compared: (i) randomly initialized, and (ii) pre-trained on ImageNet~\cite{deng2009imagenet} dataset. Note that if the network was pre-trained on ImageNet \cite{deng2009imagenet} dataset before contrastive learning, the loss for the first few epochs are relatively low, but the converged loss is quite close to the randomly initialized network. This shows that our method does not rely on pre-trained weights, and is capable of learning from scratch.

We can see that once training performance saturates and the curves flatten, \textit{e.g.}, at epoch $60$, the performance can be further improved by decaying the learning rate. We decay the learning rate twice at epochs $60$ and $90$ for the randomly initialized network, and at epochs $30$ and $60$ for the network pre-trained on ImageNet.

To provide a more intuitive understanding of the AP results in Table $2$ of the main paper, we show in Figure \ref{mn_pr_curve} the Precision Recall (PR)-curves on MovieNet test set after performing supervised learning on MovieNet training set.

\paragraph{b. Audio Modality:} As the video files in the MovieNet dataset \cite{huang2020movienet} are not yet released, we only had access to the keyframes and the pre-computed audio features for each shot in the dataset. Concatenating our $1$ keyframe-based visual shot-features with the provided pre-computed audio shot-features for scene boundary detection resulted in only marginal AP improvement from $52.34$ to $52.47$. This is in contrast to our results on the AdCuepoints dataset where incorporating learned audio features along with the learned visual features resulted in substantive AP improvement (see Table $5$ of the main paper). This observation highlights the importance of using raw audio to learn audio features for scene boundary detection, and suggests that having access to raw audio for MovieNet data could further improve the results of our approach on MovieNet dataset.

\subsubsection{Positive Key Selection}

During contrastive learning, we used ImageNet space to select our initial set of positive keys while our encoder weights were randomly initialized. This allowed us to exploit the underlying film-production process and select the right set of neighborhood shots that could offer informative data augmentation required to learn an effective embedding space.

Since the computational cost of updating the positive key set is high, we only updated the positive key set occasionally (at epochs $20$ and $50$) during training. Each update can be viewed as a re-initialization step such that between consecutive re-initializations, the argmax operation used to select positive keys does not impact differentiability.

To further distill the importance of using ImageNet space in particular for initial positive key set selection, we compare how positive key sets evolve over the course of training when they are initially selected using: (a) ImageNet space, versus (b) randomly generated space. Results from this comparison are given in Table \ref{rand_init}, and explained below.

Let $\textrm{F}_\textrm{imgnet}$ denote the feature-space learned using initial positive key-set obtained from ImageNet space, and $\textrm{K}_\textrm{imgnet}$ denote the positive key-set at each training epoch. Similarly, use $\textrm{F}_\textrm{random}$ to denote the feature-space of randomly initialized encoder, and $\textrm{K}_\textrm{random}$ as the set of positive keys at each training epoch. Table A shows the percent overlap between $\textrm{K}_\textrm{imgnet}$ and $\textrm{K}_\textrm{random}$ at the end of different epochs. We make two key observations for the results in Table A.

First, even at epoch $0$, the overlap between $\textrm{K}_\textrm{imgnet}$ and $\textrm{K}_\textrm{random}$ is already $32.12\%$, which is significantly larger than random chance (recall that our context is $16$ shots long, making the probability of there being an overlap between the two sets by random chance to be $1/16$ = $6.25\%$). 

Second, as training goes on the overlap between $\textrm{K}_\textrm{imgnet}$ and $\textrm{K}_\textrm{random}$ converges to $\sim$$75\%$ by $100$ epochs. Using this learned space for scene boundary detection task based on one keyframe produces an AP of $51.53\%$ on the MovieNet dataset.

\begin{table}[!t]
	\begin{center}
		\begin{small}
			\setlength\tabcolsep{5.5pt}
			\begin{tabu}{c|cccccc}
				\hline
				epoch \#	 &$0$	  &$20$	  &$40$	  &$60$	  &$80$	  &$100$\\\hline
				overlap \%	&32.12	&70.24	&74.42	&76.83	&75.83	&75.11\\\hline
			\end{tabu}
		\end{small}
	\end{center}
	\vspace{-0.4cm}
	\caption{\small{Overlap between $\textrm{K}_\textrm{imgnet}$ and $\textrm{K}_\textrm{random}$ for different epochs.}}
	\vspace{-0.4cm}
	\label{rand_init}
\end{table}

These observations lead us to believe that our approach stays stable so long as the feature-space used to find initial positive key-set is good enough. While ImageNet space worked out well for us, it is not a crucial requirement for the stability of our approach. In fact, even the space from randomly initialized encoder can work for our approach.

\subsection{AdCuepoints}

\noindent This section corresponds to $\S4.3.2$ of the main paper.

\subsubsection{Visual Modality}

\paragraph{a. Contrastive Learning:}

For the visual modality of AdCuepoints dataset, we used the same settings as mentioned above in $\S$~\ref{contrast_vis}-a.

\paragraph{b. Supervised Learning:}

For training the MLP for the ad cuepoint detection task using the visual modality of the AdCuepoints data, we used a fixed learning rate of $1.0$, dropout value of $0.8$, batch size of $1024$, and maximum epoch number of $200$.

\subsubsection{Audio Modality}

\paragraph{a. Contrastive Learning:}
We used the Wavegram-Logmel-CNN$14$ variant of PANNs \cite{kong2020panns} as our audio encoder. The architecture of this encoder is similar to VGG \cite{simonyan2014very} but adapted for audio. The network uses a combined wavegram and log-mel spectrogram representation. The wavegram representation is learned by expanding the time-domain input waveform to include a third dimension and convolving over this expanded input. This extra axis is analogous to frequency and allows the network to learn a joint time-frequency representation.

During contrastive learning, the weights $\theta_q$ of query encoder were randomly initialized, and the weights $\theta_k$ of key encoder were initialized to the same values as $\theta_q$. The query encoder was trained using Adam optimizer with a mini-batch size of $128$, betas of $0.9$ and $0.999$, epsilon of $1e-08$, and no weight decay. The learning rate was initialized to $0.0005$ and decayed using cosine annealing schedule\cite{kong2020panns}.

\paragraph{b. Supervised Learning:}
For training the MLP for ad cuepoint detection task on the AdCuepoints dataset for audio modality, we used a fixed learning rate of $0.01$, batch size of $512$, no dropout and maximum epoch number of $100$.

\begin{figure}[!t]
	\centering
	\includegraphics[width=0.35\textwidth]{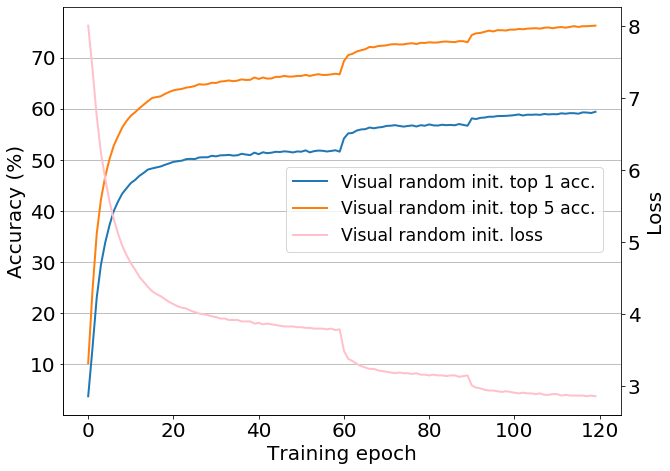}
	\vskip -0.05in
	\caption{Training curves of visual contrastive learning on AdCuepoints dataset.}
	\label{ac_tr_curve_v}
\end{figure}

\begin{figure}[!t]
	\centering
	\includegraphics[width=0.35\textwidth]{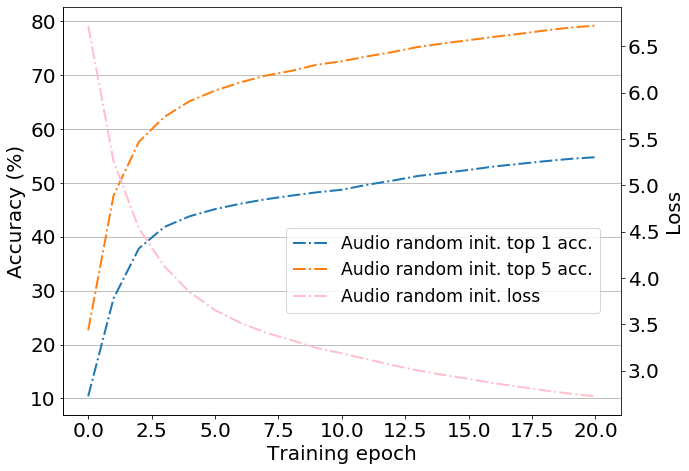}
	\vskip -0.05in
	\caption{Training curves of audio contrastive learning on AdCuepoints dataset.}
	\label{ac_tr_curve_a}
	\vskip -0.05in
\end{figure}

\begin{figure}[!t]
	\centering
	\includegraphics[width=0.4\textwidth]{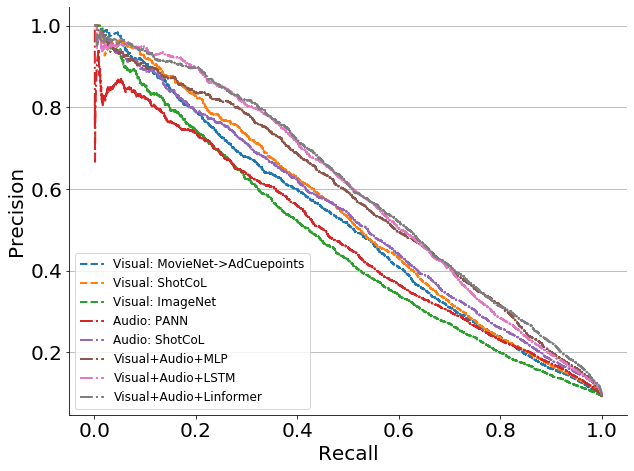}
	\vskip -0.05in
	\caption{PR-curves of test set on AdCuepoints dataset.}
	\label{ac_pr_curve}
	\vskip -0.1in
\end{figure}

\begin{figure*}[!t]
	\centering
	\includegraphics[width=1.0\textwidth]{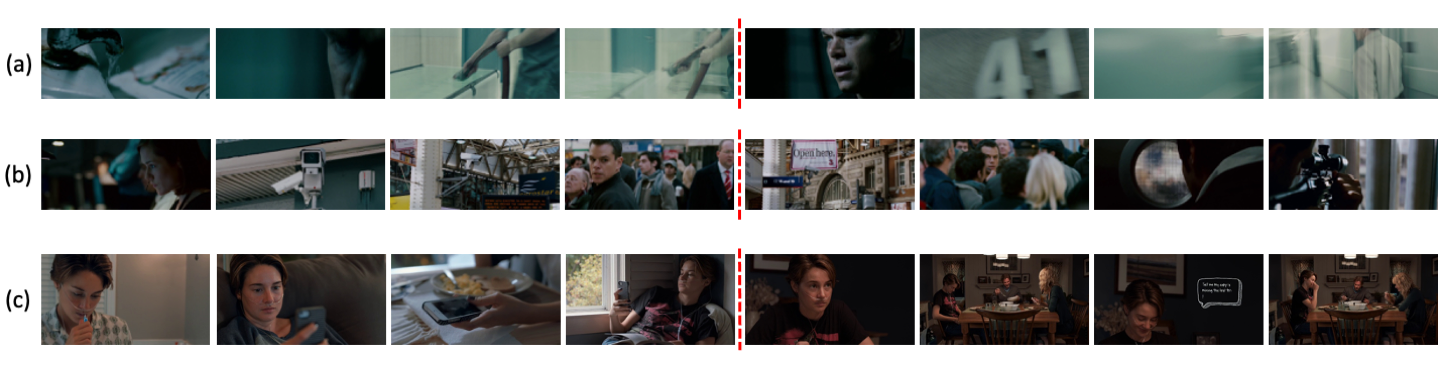}
	\vskip -0.1in
	\caption{Examples in the test set of MovieNet. Labeled scene boundaries are shown with red dashed lines.}
	\label{eg_sb}
	\vskip 0.08in
\end{figure*}

\begin{figure*}[!t]
	\centering
	\includegraphics[width=1.0\textwidth]{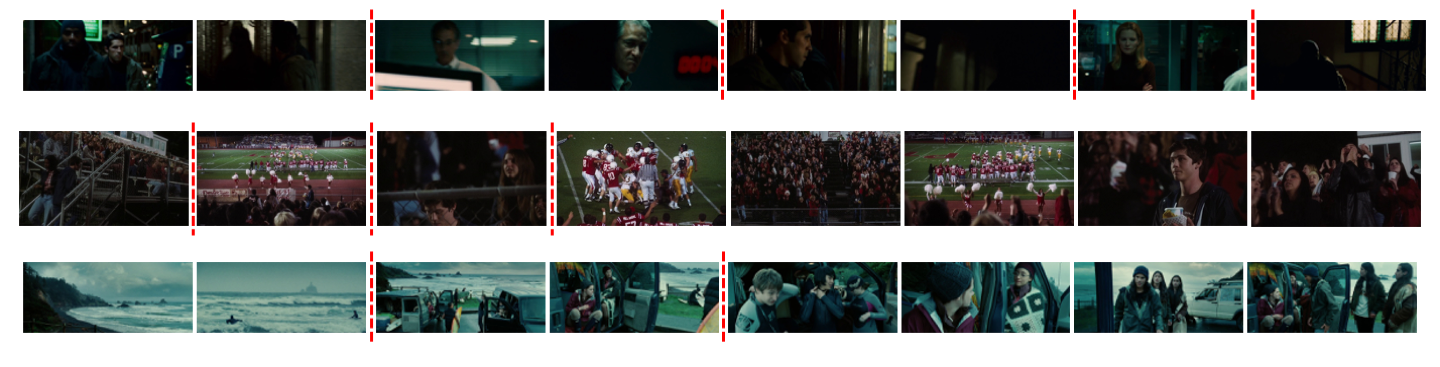}
	\vskip -0.1in
	\caption{Labeled scene boundaries in MovieNet are shown with red dashed lines.}
	\label{eg_mn}
	\vskip 0.08in
\end{figure*}

\begin{figure*}[!t]
	\centering
	\includegraphics[width=1.0\textwidth]{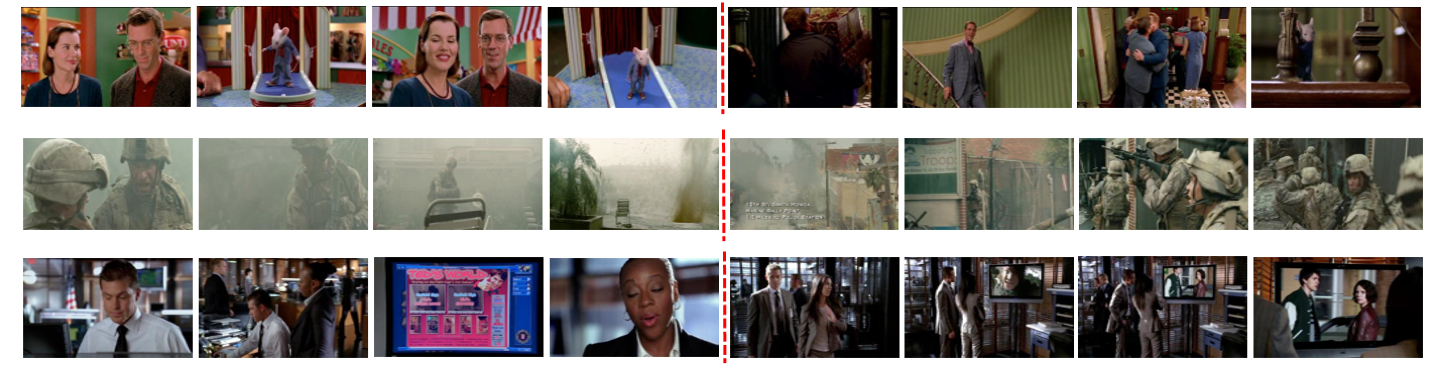}
	\vskip -0.1in
	\caption{Labeled ad cuepoints in AdCuepoints are shown with red dashed lines.}
	\label{eg_ac}
	\vskip 0.08in
\end{figure*}

\subsubsection{Audio-Visual Fusion}

\noindent In Table $5$ of the main paper, we discussed the use of more sophisticated temporal models, \textit{i.e.} Bi-LSTM and Transformer, to jointly incorporate the visual and audio features. To do this, we added an FC layer before the temporal models to map the $4096$-dimensional feature vector (audio + visual) to a more compact $2048$-dimensional feature vector. The features were then passed through the temporal model, followed by MLP for prediction. For the experiments given in Table $5$ of the main paper, each shot in the sample was treated as a separate time-step, so the sequence length was equivalent to the number of shots.

The Bi-LSTM was implemented using the LSTM module in PyTorch \cite{paszke2019pytorch}. It had two layers with $2048$ hidden units and dropout of $0.2$ between them.
We used Linformer \cite{wang2020linformer} as our choice of Transformer, which implemented sparse self-attention with linear complexity, allowing much faster runtime and lower memory usage. There were $4$ attention layers with $8$ attention heads in each layer in the Linformer, and the projection dimension was $256$.
We prepended a classification token to the beginning of each sequence, and the final hidden state of this token was used as the sequence representation for downstream classification.
 The input features were reshaped to $batch\_size \times num\_shots \times feature\_len$, to treat features from adjacent shots in the input as separate timesteps before being passed to the Linformer. The output of Linformer is then flattened to $batch\_size \times num\_shots \boldsymbol{\cdot} feature\_len$ to concatenate features from adjacent shots before being passed to the MLP. The same process is followed for the Bi-LSTM in order to treat features from adjacent shots as separate timesteps.

\subsubsection{Additional Results}

The training curves using the visual and audio modalities for contrastive learning on the AdCuepoints dataset are presented in Figure \ref{ac_tr_curve_v} and Figure \ref{ac_tr_curve_a} respectively. Note that the audio network converges notably faster than the visual network, which is consistent with the observations in \cite{xiao2020audiovisual}.
In our case, this effect is exaggerated due to the use of Adam optimizer when training the audio encoder.

To provide a more intuitive understanding of the AP results in Table $3$, $4$, and $5$ of the main paper, we show in Figure \ref{ac_pr_curve} the PR-curves on AdCuepoints test set after performing supervised learning on AdCuepoints training set.

Recall that for the visual modality of the MovieNet, we showed the effectiveness of using our proposed shot-similarity compared to existing image augmentation schemes (Table $2$, row $8$ and row 11 in the main paper). Along similar lines, we explored the effectiveness of using existing audio augmentation schemes (\textit{e.g.} SpecAugment \cite{Park2019}) compared to our proposed audio-shot similarity for the AdCuepoints dataset.
Contrasting a query shot with its augmented version (using SpecAugment~\cite{Park2019}) during contrastive learning, we can achieve an AP of $50.53$ using the $10$ shot setting. Our shot similarity-learning approach instead can achieve an AP of $53.27$. This result further validates applicability of our approach on audio modality.

\section{Qualitative Results}

\subsection{Challenging Examples}

\noindent To give an intuitive sense of the difficulty level of the scene boundary detection task, we show in Figure \ref{eg_sb} a few examples of the labeled scene boundaries in the test set of MovieNet data. As can be observed, even humans can have difficulty confidently disambiguating whether the shot boundaries in Figure \ref{eg_sb} are scene boundaries or not.

\subsection{MovieNet vs AdCuepoints}

\noindent Recall that ad cuepoints are a special case of scene boundaries. To provide more intuition behind this point, in Figure \ref{eg_mn} and Figure \ref{eg_ac} we present some representative examples that demonstrate the differences between the MovieNet and AdCuepoints datasets. As can be observed in Figure \ref{eg_mn}, scene boundaries can be semantically quite close to each other, and arbitrarily inserting video ads to such scene boundaries can break the flow of the storyline and interrupt the viewing experience of the audience.

In contrast, as shown in Figure \ref{eg_ac} ad cuepoints are more distinguishable and isolated from each other. Therefore, inserting ads at such points is likely to result in minimal disruption of the storyline as the scenes before and after the ad cuepoints are more distinguishable and semantically unrelated from each other.

\subsection{Additional Nearest Neighbor Results}

\begin{figure*}[!t]
	\centering
	\includegraphics[width=0.98\textwidth]{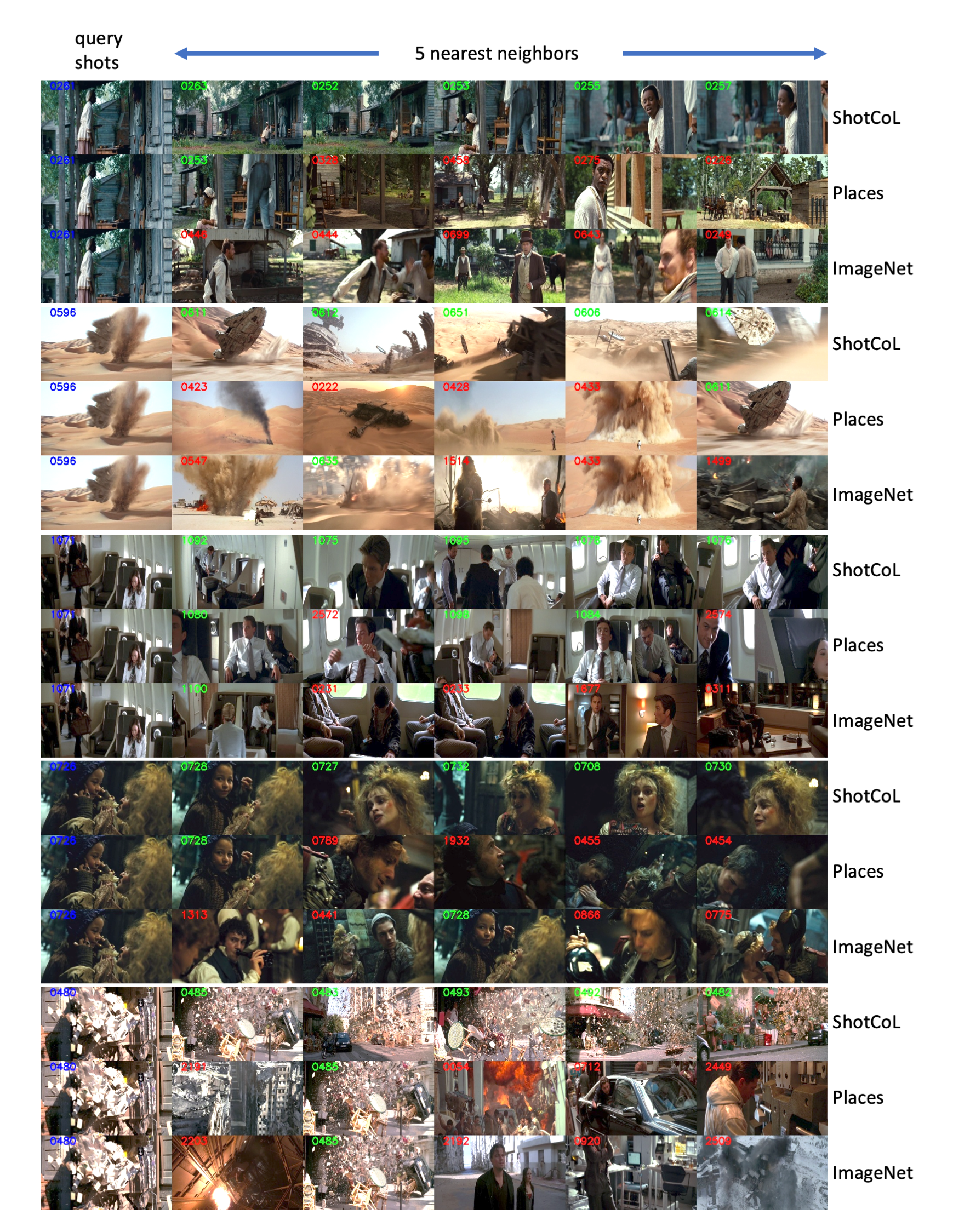}
	\caption{Additional nearest neighbor results. Shot indices are displayed at top-left corners where blue indicates query shot, green indicates shot from the same scene as query, and red indicates shot from a different scene.}
	\label{nn_eg}
\end{figure*}

\noindent Recall that we discussed the effectiveness of our learned shot representation for the task of scene boundary detection in $\S4.1$ in the main paper.
In Figure \ref{nn_eg} of this document, we provide some additional results to underscore this point. We present $5$ nearest neighbor shots retrieved for a query shot using different shot-representations. We compare our learned representation with ImageNet feature \cite{deng2009imagenet} and Places feature \cite{zhou2017places} to tell whether the $5$ nearest neighbor shots are from the same scene or not. As can be observed, while results retrieved using Places and ImageNet features are visually quite similar to the query-shot, almost none of them are from the query-shot’s scene (indicated on the top left of shot-frame). In contrast, results from our shot-representation are all from the same scene even though the appearance of the retrieved shots does not exactly match query shot. This shows that our learned shot-representation is able to effectively encode the local scene-structure.

{\small
\bibliographystyle{ieee_fullname}
\bibliography{ref2}
}